# UAV-Based Environmental Monitoring of Rip-Current Indicators Using Wavelet-Derived Texture Features

Yonatan Ben Avraham[1] (Yonatan.Ben.avraham@s.afeka.ac.il)
Baruch Binyaminov[1] (Baruch.Binyamniov@s.afeka.ac.il)
Yehudit Aperstein[1,*] (apersteiny@afeka.ac.il)

[1]Intelligent Systems, Afeka Academic College of Engineering, Tel Aviv 6998812, Israel
*Corresponding author: apersteiny@afeka.ac.il

## Abstract

Rip currents are recurrent coastal natural hazards that threaten beachgoers and create operational challenges for lifeguards and coastal managers. Reliable monitoring from standard RGB (red-green-blue) imagery acquired by unmanned aerial vehicles (UAVs) remains difficult because hazardous channels often appear as subtle gaps in breaking waves, foam texture, or sediment patterns, and these signatures are affected by illumination, sea state, and environmental noise. This study presents a physically informed coastal environmental monitoring workflow for detecting visually expressed rip-current indicators that integrates wavelet-derived spatial-frequency texture features with deep learning. We evaluate multiple strategies for incorporating Discrete Wavelet Transform features into convolutional architectures, from computationally efficient channel replacement to dual-stream fusion with attention mechanisms. Performance is assessed against a standard RGB baseline using a task-specific convolutional neural network for image-level presence classification and a YOLOv8 model for object-level localization. Under the evaluated dataset conditions, integrating wavelet-derived texture features improves performance over RGB-only models. The dual-stream architecture achieves the strongest classification performance, exceeding 95% accuracy with high recall, while channel replacement is most effective for YOLOv8 object detection, reaching 94% mAP@50 for localization. Explainable artificial intelligence analyses provide qualitative evidence that the models attend to visually plausible wave-gap regions associated with rip currents. These results suggest that under the conditions of the evaluated dataset, physically informed wavelet integration may support UAV-based decision-support tools for interpretable beach-safety risk mitigation.

# 1. Introduction

Rip currents are powerful, narrow channels of fast-moving water prevalent along beaches worldwide. They represent one of the most hazardous coastal phenomena, accounting for more than 80% of surf-beach rescues and contributing to a substantial number of drowning fatalities annually (NOAA 2024; USLA n.d.; Brewster et al. 2019). Public awareness of this hazard remains incomplete; O'Halloran and Silver (2025) found that only 44% of surveyed American adults recognized rip currents as the major cause of drownings in the United States. Despite their significant danger, rip currents remain notoriously difficult to detect visually, especially for untrained beachgoers. Irvine et al. (2025) reported that beach users correctly identified rip currents only about half of the time when assessed using photographs and videos. Rip currents often lack clear surface signatures, appearing deceptively calm compared to the surrounding breaking waves (Maryan et al. 2019). This visual ambiguity is particularly critical on crowded public beaches, where timely and accurate identification of these hazards can directly influence safety operations and prevent loss of life. The long-standing need for automated visual monitoring is also reflected in early image-based detection systems, including the automated rip-current detection approach proposed by Perrier (2005).

Historically, coastal monitoring relied on in-situ measurements using Acoustic Doppler Current Profilers (ADCPs) or GPS drifters (Austin et al. 2013). While these methods provide high-accuracy flow data, they suffer from limited spatial coverage, high deployment costs, and are impractical for real-time public monitoring. Consequently, the field has shifted toward remote sensing techniques. Traditional optical systems, such as time-averaged photography (TIMEX), have been widely used to visualize breaking wave patterns (Holman and Stanley 2007). However, these methods typically rely on manual interpretation or handcrafted heuristics, making them sensitive to variable lighting conditions, camera angles, and local beach morphology.

Within a coastal environmental monitoring context, UAV imagery offers a practical way to observe recurrent hazardous surf-zone conditions at the beach scale. Low-altitude aerial observations can repeatedly capture the surf zone at spatial resolutions relevant to rip-current morphology, while avoiding the cost and limited coverage of fixed in-situ instrumentation. For coastal managers and lifeguard services, the central challenge is therefore early identification of hazardous surf-zone conditions and translation of ambiguous visual cues into information that can support beach-safety risk mitigation.

In this setting, deep learning is used as an enabling component within a physically informed hazard-monitoring workflow rather than as the central novelty by itself. Convolutional Neural Networks (CNNs) and detectors like Faster R-CNN and YOLO have demonstrated promising results in automating rip current detection (De Silva et al. 2021; Dumitriu et al. 2023). Nevertheless, standard models operating on RGB imagery face important reliability limitations for coastal environmental monitoring. They often struggle to distinguish the subtle texture of a rip channel from the noise of breaking waves, leading to false negatives in complex sea states. Furthermore, deep models are frequently criticized for being "black boxes," lacking the operational interpretability required for safety-critical beach management (Rampal et al. 2022).

To address the limitations of RGB-only analysis, recent research has revisited physically interpretable signal-processing techniques that are well suited to remote sensing of coastal hydrodynamics. Notably, Wang et al. (2025) demonstrated that the Discrete Wavelet Transform (DWT) can effectively extract hydrodynamic features by highlighting the texture gradients between high-energy breaking waves and the lower-energy rip channel. While their work established the physical utility of wavelets, it relied on traditional convolution and thresholding, which may lack the generalization capabilities and speed of modern data-driven architecture.

This study is framed as a coastal environmental monitoring workflow for the early identification of hazardous surf-zone conditions. Its contribution is to translate physically meaningful wavelet-derived hydrodynamic textures into operational rip-current hazard detection and localization products that can support beach-safety risk mitigation. We introduce and compare distinct spectral-spatial integration approaches, including a computation-efficient channel-replacement strategy, where the red channel is substituted with a DWT-based energy map, and a dual-stream architecture that processes spatial and spectral features in parallel. The primary aim is not to present artificial-intelligence novelty alone, but to evaluate whether physically informed representations make UAV imagery more informative, interpretable, and reliable for recurrent coastal hazard monitoring. Our results show improved precision and recall under the evaluated dataset conditions, supporting the potential use of spectral-spatial representations in UAV-based beach-safety workflows. In this study, the term spectral-spatial refers to wavelet-derived spatial-frequency texture representations extracted from RGB imagery, not to multispectral or hyperspectral sensing.

Furthermore, we employ Explainable AI (XAI) techniques to examine whether model attention aligns with physically meaningful wave-gap regions rather than obvious environmental artifacts.

The main contributions of this study are as follows:

1. Physically informed coastal hazard-monitoring workflow: We propose a UAV-based workflow that embeds wavelet-derived hydrodynamic texture information into rip-current hazard detection, supporting the screening and prioritization of candidate hazardous surf-zone regions from standard RGB imagery.
2. Spectral-spatial representations for beach-safety risk mitigation: We evaluate multiple ways of integrating wavelet information into learning models, including wavelet-only inputs, early RGB-wavelet fusion, image fusion, hue-saturation-wavelet representation, channel replacement, and dual-stream RGB-wavelet fusion, to determine which representations are most useful for operational hazard monitoring.
3. Complementary hazard detection and localization evaluation: We assess the proposed representations in both image-level rip-current hazard classification and object-level localization, showing that dual-stream fusion is most effective for classification, whereas channel replacement provides the strongest YOLOv8 localization performance.
4. Qualitative interpretability and failure-mode analysis: We use Grad-CAM++, EigenCAM, and qualitative error analysis to examine whether model activations coincide with visually plausible surf-zone structures and to identify failure modes such as bathymetric mimicry and subtle narrow rip-current geometry.

The remainder of the paper is organized as follows. Section 2 reviews related work on rip-current monitoring, coastal hazard detection, deep learning, and wavelet-based image analysis. Section 3 presents the coastal monitoring and signal-processing background underlying the proposed wavelet representation. Section 4 describes the dataset, preprocessing pipeline, channel-replacement strategy, dual-stream architecture, and comparative fusion methods. Section 5 details the experimental setup and evaluation metrics. Section 6 reports the classification, detection, ablation, and explainability results. Section 7 discusses the implications of the findings for UAV-based coastal environmental monitoring, including operational interpretability, beach-safety risk mitigation, and reliability limitations. Section 8 summarizes limitations and future research directions.

## 2. Literature Review

Rip current research spans coastal geomorphology, optical physics, and computer vision. Early studies established the foundational understanding of rip morphology and circulation patterns. Austin et al. (2013) and recent critical reviews (Valipour 2025) provide comprehensive overviews of the physical mechanisms driving rip channels, outlining key visual indicators-such as darker gaps in breaking waves or seaward-moving sediment plumes-that serve as the ground truth for automated detection. Complementing empirical observation, numerical approaches like Boussinesq modeling (Liu et al. 2025) have been used to simulate rip occurrences on featureless beaches, though such hydrodynamic models are often computationally intensive for real-time monitoring.

Crucial to visual detection is the understanding of how light interacts with water. Foundational optical studies by Pope and Fry (1997) demonstrated that water absorption increases strongly toward longer wavelengths; consequently, red light attenuates significantly faster than blue or green light in aquatic environments. This optical property suggests that red channels in RGB imagery may contain less reliable structural information regarding underwater or near-surface features. Conversely, blue and blue-green spectral information is often useful in aquatic imaging contexts, including dye-tracking studies of rip-current dynamics (Kim and Kim 2021). This physical rationale motivates our investigation of strategies in which the red channel is replaced by a physically meaningful wavelet-energy descriptor in coastal UAV imagery. Green and blue channels were therefore retained because they generally preserve more aquatic visual structure, whereas the red channel is expected to be less informative in water-imaging contexts.

The transition to automated detection began with rule-based heuristics. An early 2005 patent (Perrier 2005) proposed using handcrafted color and texture cues, marking early interest in image-driven analysis. Systematic academic attempts emerged with Maryan et al. (2019) who demonstrated that even shallow Convolutional Neural Networks (CNNs) outperform classical feature-based models like SVMs. Subsequent deep learning advancements focused on temporal dynamics and interpretability. De Silva et al. (2021) introduced a Faster R-CNN framework with temporal aggregation to stabilize detections in dynamic surf, while RipViz (De Silva et al. 2023) incorporated optical flow and LSTM autoencoders to model water movement. To address the "black box" nature of these models, Rampal et al. (2022) applied Explainable AI (XAI) techniques, using Grad-CAM to visualize the spatial regions driving predictions-a critical step for operational trust.

Recent literature emphasizes the deployment of lightweight models on Unmanned Aerial Vehicles (UAVs) and edge devices. RipFinder (Khan et al. 2025a) and RipScout (Khan et al. 2025b) presented mobile-ready frameworks for real-time detection, highlighting the shift towards accessible coastal surveillance. Building on this trend, Kumar et al. (2024) proposed an integrative early warning system that deploys optimized object detection models directly on edge devices, enabling real-time hazard identification without relying on heavy shore-based computing. Similarly, Ali et al. (2025) explored IBN-driven analysis using UAVs for next-generation surveillance. The field is currently advancing towards precise segmentation and optimization of modern architecture. The 2025 RipSeg Challenge (Dumitriu et al. 2025) and Dumitriu et al. (2023) established benchmarks for dense segmentation using YOLOv8, enabling the mapping of exact rip contours. Furthermore, recent studies on small datasets (Putri et al. 2025) have investigated optimizing data augmentation parameters specifically for rip current detection, addressing the scarcity of labeled coastal data.

Parallel to deep learning, wavelet-based methods remain highly relevant for their computational efficiency and ability to capture physical textures. Wang et al. (2025) utilized the Discrete Wavelet Transform (DWT) to extract directional water-flow features, effectively enhancing breaker fronts while suppressing background noise. Their work highlights that wavelets can capture hydrodynamic gradients that pure RGB models might miss. Demonstrating the broader applicability of this paradigm in maritime environments, Zhou et al. (2025) recently showed that a dual-branch Wavelet-CNN architecture significantly improves the detection of small objects in noisy sea states by decoupling structural high-frequency details from complex background waves.

This study synthesizes these domains by linking coastal optical physics, UAV remote sensing, and operational hazard detection. While Wang et al. (2025) demonstrated the feature extraction capabilities of wavelets, their integration into end-to-end deep neural networks remains underexplored. Building on the optical principles regarding channel information (Pope and Fry 1997), we investigate optimal strategies for fusing spectral and spatial data. We propose and compare a computation-efficient "Channel Replacement" strategy against robust "Dual-Stream" architecture equipped with attention mechanisms. By embedding physical texture features into both task-specific classifiers and advanced object detectors (YOLOv8), we aim to improve the remote-sensing utility of UAV imagery for rip-current monitoring while evaluating the trade-offs required for operational beach-safety deployment.

## 3. Background

This section establishes the coastal hazard-monitoring and signal-processing basis for the proposed rip-current monitoring framework. In UAV observations of the surf zone, rip currents are rarely expressed as simple color anomalies; they are more often manifested through spatial discontinuities in breaking-wave patterns, foam texture, sediment plumes, and the transition between turbulent and relatively quiescent water. The role of the background presented here is therefore to connect the physical appearance of rip currents in aerial imagery with spectral-spatial image representations that can be exploited by lightweight operational models.

### 3.1 Discrete Wavelet Transform (DWT)

The Discrete Wavelet Transform (DWT) is particularly relevant for coastal remote sensing because nearshore image texture is non-stationary. Wave breaking, foam bands, rip channels, and shore-parallel surf structures vary over short spatial distances and at multiple scales. Unlike a global Fourier representation, which describes frequency content without preserving where that content occurs, DWT provides a localized scale-space decomposition. This property makes it suitable for separating broad intensity structure from localized hydrodynamic texture gradients that may indicate rip-current boundaries.

For two-dimensional UAV imagery, DWT is applied separably along image rows and columns. Each decomposition level produces four sub-bands that encode complementary spatial-frequency information:

1. Approximation (LL): Low-frequency image structure, including broad illumination gradients, water-color patterns, and large-scale surf-zone organization.
2. Horizontal Detail (LH): Directional texture changes associated with cross-shore or shore-parallel wave transitions, depending on image orientation.
3. Vertical Detail (HL): Orthogonal edge responses that can capture narrow foam discontinuities, channel margins, and localized surf-line breaks.
4. Diagonal Detail (HH): High-frequency diagonal texture and fine-scale turbulence, including foam fragmentation and small-scale wave-front irregularity.

In this applied remote-sensing study, we focus on the spectral descriptor extracted from each UAV image rather than on the full filter-bank derivation of DWT. After a first-level two-dimensional DWT, the high-frequency detail sub-bands capture localized horizontal, vertical, and diagonal texture changes associated with foam gradients, wave-front discontinuities, and rip-channel boundaries. We therefore define the wavelet-energy map used in the Channel Replacement and fusion experiments as:

$$W(x,y) = \mathrm{Norm}(\mid LH_1(x,y) \mid + \mid HL_1(x,y) \mid + \mid HH_1(x,y) \mid) \quad (1)$$

where $LH_1$, $HL_1$, and $HH_1$are the first-level horizontal, vertical, and diagonal detail sub-bands, respectively, and $\mathrm{Norm}(\cdot)$denotes normalization to the image-intensity range used by the learning models. Equation (1) provides the physically interpretable spectral layer $W$that replaces or complements RGB information in the proposed UAV-based rip-current monitoring workflow.

Recursive decomposition of the LL sub-band produces a multi-resolution representation, allowing coarse surf-zone organization and fine hydrodynamic texture to be examined within a common framework. For rip-current monitoring, this is valuable because the hazard may appear simultaneously as a broad gap in breaking waves and as fine localized texture at the turbulent boundary between the rip channel and adjacent breakers.

From a remote-sensing perspective, the wavelet sub-bands should not be interpreted merely as abstract image features. They provide a compact representation of physically meaningful surf-zone signals, including abrupt foam gradients, suppressed breaking inside rip channels, sediment-laden streaks, and transitions between energetic and low-energy water surfaces. Embedding these descriptors into a learning model therefore gives the model access to hydrodynamic texture cues that are often weak or ambiguous in raw RGB intensity values.

### 3.2 Choice of Wavelet Basis

The selection of a wavelet basis affects how localized coastal textures are represented. This study focuses on the Daubechies family because these wavelets combine compact support with multiple vanishing moments, enabling localized edge and texture detection while retaining sensitivity to short spatial transitions. Such properties are appropriate for UAV rip-current imagery, where the relevant signal is often a narrow and irregular transition between foam-rich breaking waves and smoother rip-channel water rather than a large, uniform spectral region.

This basis also supports the operational objective of the study. Daubechies wavelets can be computed efficiently, generate compact feature maps, and preserve spatial localization sufficiently for integration with lightweight classifiers and object detectors. The resulting representation is therefore compatible with the goal of developing interpretable, UAV-based coastal natural-hazard monitoring tools for beach-safety risk mitigation and operationally meaningful hazard detection.

## 4. Methodology

This section outlines the comprehensive methodology employed to detect and classify rip currents. It details the dataset composition and preprocessing techniques, followed by an in-depth description of the architectural designs for both the classification networks and the object detection models.

### 4.1 Dataset

To ensure consistent benchmarking and reproducibility, this study uses the publicly available Rip Current Monitoring dataset, Version 1, from Roboflow Universe (Rip Currents 2025). The dataset contains 2,246 aerial coastal images acquired under diverse nearshore conditions, including variation in water turbidity, illumination, surf-zone texture, and sea state. The dataset includes YOLO-format bounding-box annotations for rip-current regions. Consistent with prior image-based rip-current detection studies, in which rip-current labels are assigned from visually identifiable surf-zone signatures such as wave-breaking gaps, darker calmer channels, sediment plumes, and texture discontinuities (De Silva et al. 2021; Zhu et al. 2022), the annotations in this dataset are treated as visual rip-current region labels for object-detection benchmarking.

The official dataset split was used for training, validation, and testing, comprising 1,566 training images, 453 validation images, and 227 test images. All images were standardized to a resolution of 320 × 320 pixels, and EXIF orientation metadata was removed. For the object detection task, the original YOLO annotations were used. For the binary classification task, image-level labels were derived from the annotations: images containing one or more rip-current bounding boxes were labeled as positive, whereas images without bounding boxes were labeled as negative. The dataset is moderately imbalanced, with approximately 69% positive samples and 31% negative samples. Therefore, in addition to accuracy, we report precision, recall, and F1-score, with particular attention to recall because missed rip-current detections are safety-critical. Representative examples of the two classes are shown in Fig. 1.

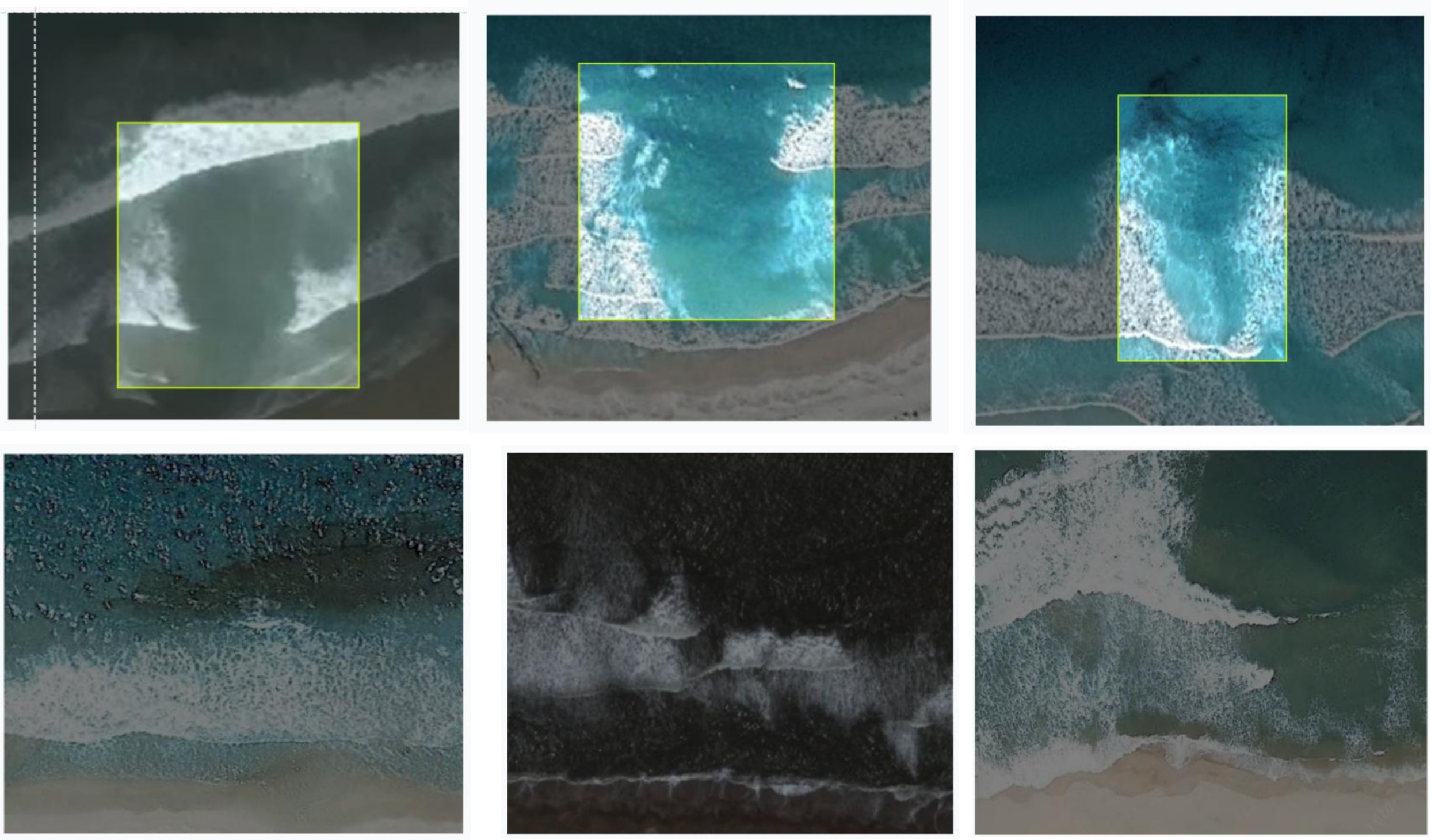

**Fig. 1.** Representative dataset samples showing rip-current images (top row) and safe breaking-wave images (bottom row). The subtle visual contrast between classes motivates physically informed spectral-spatial feature extraction.

*Data leakage prevention.* Coastal drone imagery is susceptible to temporal and spatial leakage, since sequential frames, burst captures, or repeated UAV passes over the same beach region can produce highly similar images across different splits. Such overlap may inflate test performance by allowing models to learn background-specific visual patterns rather than general rip-current indicators. To assess this risk for the final held-out evaluation, we performed a near-duplicate analysis between the training and test splits using perceptual hashing (pHash). For each training and test image, a structural hash signature was computed, and image pairs were compared using Hamming distance. No functionally identical image pairs, defined as Hamming distance $\leq 5$, were found between the training and test sets. This check supports the use of the held-out test set as an independent evaluation set and reduces the likelihood that reported performance reflects background memorization.

### 4.2 Data Preprocessing and Augmentation

Given the distinct requirements of binary classification and object detection, the dataset was processed through two separate pipelines to ensure optimal model performance.

#### **4.2.1** *Preprocessing for Binary Classification*

To adapt the raw dataset for the task-specific convolutional neural network architectures, the following systematic pipeline was applied:

1. Label Conversion: Image-level labels were derived from the YOLO annotations. Images containing one or more bounding boxes were assigned a positive label (Rip Current), while images devoid of annotations were assigned a negative label.
2. Spatial Cropping: To prevent the models from learning spurious correlations associated with the shoreline (e.g., sand color, beach structures), a deterministic bottom crop was applied, removing the lower 20% of each image. This focuses the network's attention purely on the surf zone and nearshore hydrodynamics.

3. Standardization: All cropped images were resized to a uniform resolution of 320×320 pixels, normalized to the range [0, 1], and EXIF orientation metadata was removed.
4. Online Data Augmentation: To improve generalization and mitigate overfitting, random horizontal flips were applied dynamically during training. This augmentation preserves the physical validity of the data by accounting for the left-right symmetry often observed in surf zone wave breaking patterns.

**4.2.2** *Preprocessing for Object Detection (YOLOv8)*

For the object detection task, maintaining the integrity of the spatial coordinates for the bounding boxes was critical. Therefore, the deterministic bottom cropping used in the classification pipeline was bypassed. Full-frame images were utilized to preserve the complete contextual field of view. The images were processed directly by the YOLOv8 data loader, which inherently handles dynamic resizing to 640×640 pixels (imgsz=640) while maintaining aspect ratios through internal padding. Standard batch normalization and native YOLO augmentations were automatically applied during the training phase to ensure robust feature learning across varying scales and orientations.

## 4.3 Physically Informed Channel Replacement Strategy

Building on the theoretical absorption properties discussed in Section 3, we implement a targeted channel replacement strategy. Since the red channel $(R)$is expected to contain less structural information about submerged or near-surface hydrodynamic features and may be more affected by surface reflection noise, we substitute it with the wavelet-energy map $(W)$derived from the high-frequency DWT sub-bands $(LH_1, HL_1, HH_1)$. Formally, given an input RGB image $I_{\text{RGB}}$, the transformed input $I_{\text{new}}$is constructed as:

$$I_{\text{new}} = [W, G, B] \qquad (2)$$

where $W$is the wavelet-energy map, and $G$and $B$are the original green and blue image channels, respectively. This transformation replaces the red channel with a physically motivated spectral-texture descriptor while preserving a standard three-channel input format.

This technique embeds spectral texture information explicitly into the input layer, allowing standard 3-channel CNN backbones (like YOLO or ResNet) to process hydrodynamic features without requiring architectural modification. The Channel Replacement workflow is summarized in Fig. 2.

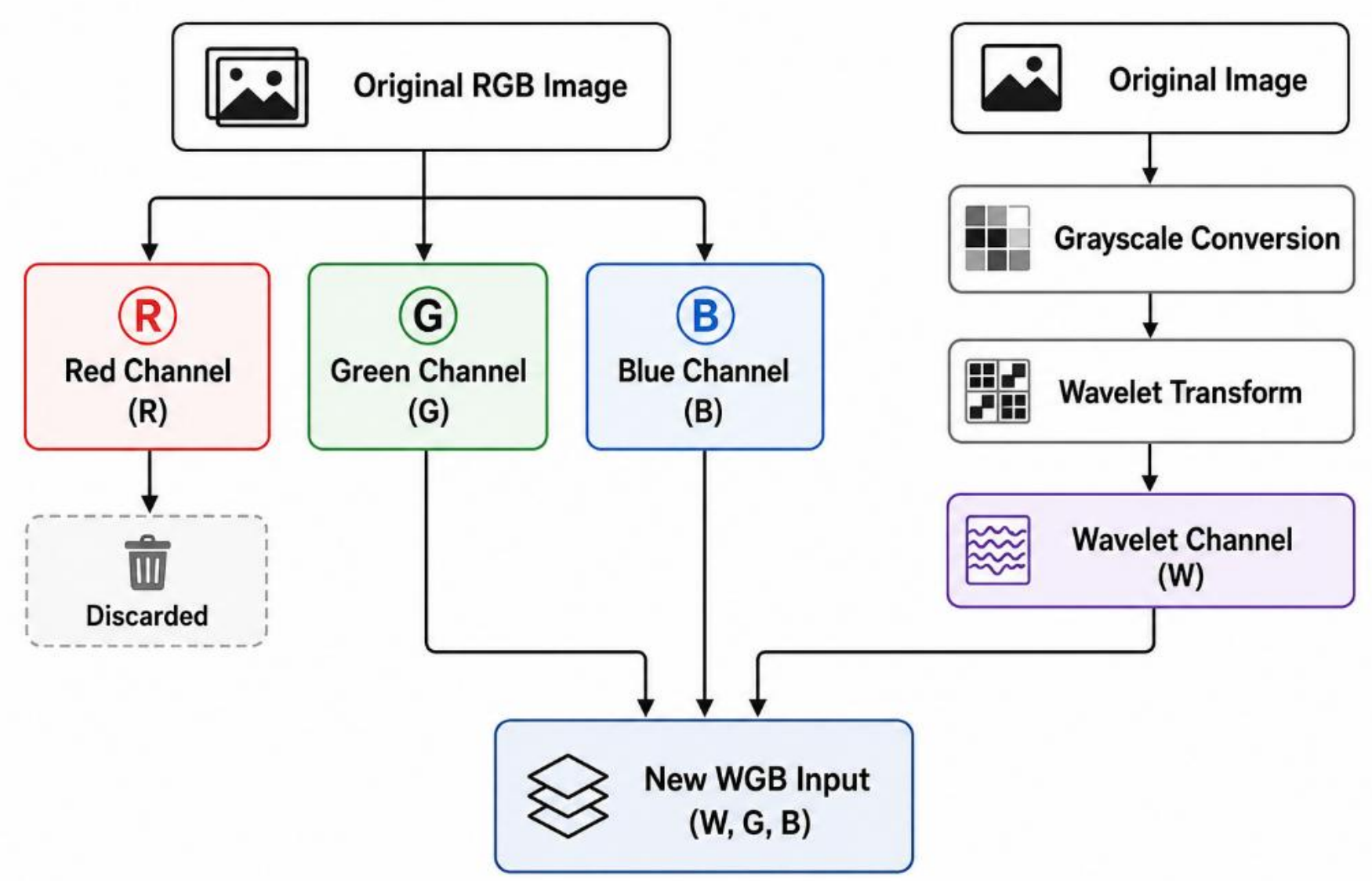

**Fig. 2.** Channel Replacement workflow. A derived wavelet energy channel (W) replaces the original red channel to create a W-G-B composite that embeds hydrodynamic texture information into a standard three-channel input.

### 4.4 Network Architectures

For Binary Classification: We developed a task-specific Dual-Stream convolutional neural network capable of fusing spatial (RGB) and spectral (Wavelet) features in parallel. Additionally, we evaluated a Single-Stream variant adapted to process the Channel-Replaced input for comparative analysis.

For Object Detection: We utilized the YOLOv8 architecture, specifically configuring its input layer to accept the Channel-Replaced composite images, thereby embedding hydrodynamic texture data directly into the detector's backbone.

#### *4.4.1 Classification Models*

The core of our classification framework is the Dual-Stream Architecture, designed to explicitly leverage both visual appearance and hydrodynamic texture. As illustrated in Fig. 3, the network comprises two parallel branches:

1. The Spatial Stream**:** A CNN branch processing the standard 3-channel RGB input to capture color and shape features.
2. The Spectral Stream: A parallel CNN branch processing the 4-channel Wavelet stack (LL, LH, HL, HH) to extract frequency-domain texture gradients.

   Through empirical ablation studies, we explored various input combinations for these streams, such as reducing the spatial stream to 2 channels (G, B) or changing the number of wavelet sub-bands. However, the comprehensive 4-channel spectral and 3-channel spatial configuration yielded the most robust feature representation.

Both streams utilize identical backbone structures consisting of convolutional layers followed by batch normalization and ReLU activation. The feature maps from both streams are fused via a concatenation layer, creating a unified representation that is subsequently fed into the Attention Module (detailed in Section 4.4.2) and a fully connected classifier.

To rigorously evaluate the effectiveness of the dual-stream fusion strategy, we implemented four distinct Single-Stream configurations for comparison:

1. RGB Baseline: A standard CNN processing only spatial information (3 channels),

representing the common approach in current literature.

2. Pure Wavelet: A network processing only spectral features (4 channels), testing the hypothesis that texture alone is sufficient for detection.
3. Early Fusion (4-Channel Stack): A model where the Wavelet energy map is concatenated with the RGB channels at the input level $(R, G, B, W)$. This tests whether performance gains stem merely from data availability or from the architectural separation.
4. Channel Replacement: A model utilizing the 3-channel composite input $(W, G, B)$ defined in Section 4.3. This configuration evaluates the efficiency of discarding the Red channel in favor of spectral data within a standard CNN architecture.

*4.4.2 Attention Mechanism (CBAM)*

To enable the networks to effectively prioritize the most discriminative features, we integrated a Convolutional Block Attention Module (CBAM) into our architectures. CBAM is a lightweight, sequential attention module that refines intermediate feature maps along two separate dimensions:

1. Channel Attention: This sub-module learns inter-channel relationships, assigning higher weights to channels containing critical rip current signatures (e.g., specific high-frequency wavelet sub-bands) while suppressing redundant noise.
2. Spatial Attention: Subsequently, the spatial sub-module highlights informative regions-such as the active surf zone and wave gaps-while masking out background noise like static sand.

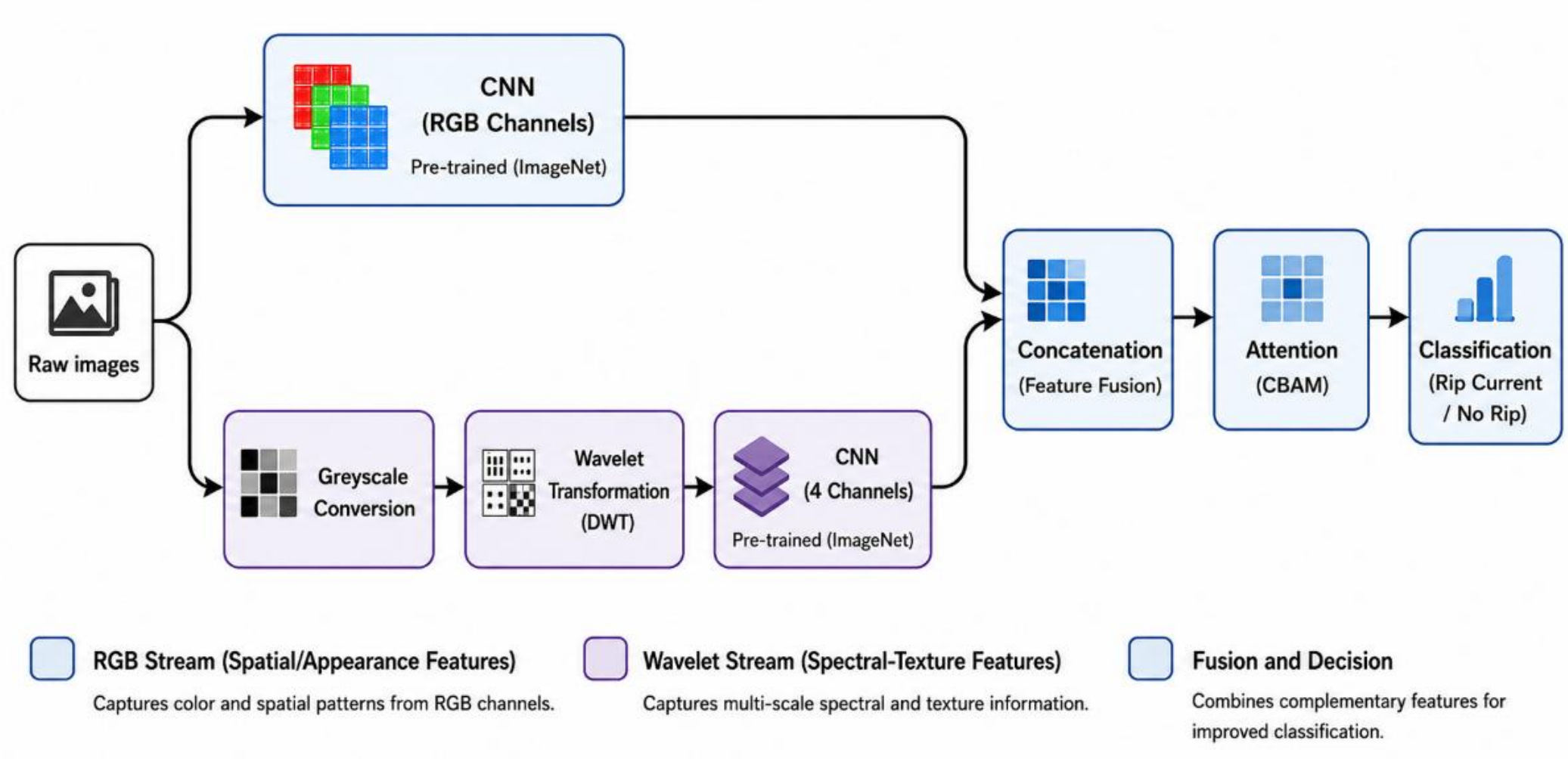

**Fig.3.** Dual-stream spectral-spatial CNN architecture for rip-current classification. The spatial branch processes RGB imagery, the spectral branch processes DWT sub-bands, and the fused representation is passed through attention and classification layers.

*4.4.3 Object detection model (YOLOv8)*

For the spatial localization of rip currents, we employed the YOLOv8 (You Only Look Once version 8) architecture, specifically the Nano variant (YOLOv8n). This model was selected due to its optimal balance between high detection accuracy and low computational overhead, making it highly suitable for real-time coastal monitoring applications. YOLOv8 utilizes an anchor-free detection head and a decoupled architecture, which significantly improves its ability to localize dynamic, irregularly shaped features such as rip currents in complex marine environments.

To seamlessly integrate the physically informed monitoring representation, the native 3-channel input layer of the YOLOv8 backbone was supplied with the Channel-Replaced composite images $(W, G, B)$. This allowed the detector to simultaneously learn spatial boundaries and spectral energy gradients without requiring architectural modifications.

### 4.5 Comparative Integration Strategies

To evaluate the effectiveness of the proposed replacement strategy, we investigated alternative fusion methods:

- HSW (Hue-Saturation-Wavelet): Instead of RGB space, we converted images to HSV and replaced the Value (V) channel-which correlates with brightness-with the Wavelet map (W), aiming to decouple color from texture intensity.
- Directional Wavelet Composite: To evaluate whether the networks could learn solely from directional frequency gradients, we constructed a purely spectral 3-channel input. The original image was converted to grayscale and subjected to a 2D Discrete Wavelet Transform (DWT). We extracted the first-level approximation $LL_1$, horizontal detail $LH_1$, and vertical detail $HL_1$ coefficients. These matrices were normalized and merged to form a pseudo-RGB composite $(LL_1, LH_1, HL_1)$ allowing the standard architecture to process isolated hydrodynamic directional features without any spatial color data.
- Image Fusion: To test a direct spatial-spectral overlap, we applied a pixel-level weighted fusion. First, a normalized Wavelet energy map $W$ was computed from the absolute sum of the first-level high-frequency sub-bands $(LH_1, HL_1, HH_1)$, consistent with Equation (1), derived from a first-level Daubechies 4 (db4) transform. This single-channel map was replicated across three channels and blended with the original RGB image $I_{\text{RGB}}$ using a linear combination: $I_{\text{fused}} = 0.7\, I_{\text{RGB}} + 0.3\, W$. This configuration superimposes the hydrodynamic texture intensity directly onto the visual color data, testing whether the network can internally separate the overlaid features.

These configurations serve as baselines to demonstrate that explicit channel substitution (Section 4.3) and dual-stream processing (Section 4.4.1) yield superior feature representation for rip current analysis.

## 5. Experimental Setup

This section describes the experimental framework used to train and evaluate the proposed models. It covers the specific implementation details, hyperparameter configurations, evaluation metrics, and the hardware environment utilized to ensure the reproducibility of the experiments.

### 5.1 Implementation Details

All experiments were conducted using the PyTorch deep learning framework with GPU acceleration to facilitate efficient computational training.

*Binary Classification Training Configuration:* The task-specific classification architectures (both Single-Stream and Dual-Stream) were trained using the Adam optimizer with a learning rate of 0.001. The network weights were optimized using the Binary Cross-Entropy (BCE) loss function. Training was executed with a batch size of 32 for a maximum of 100 epochs. To actively mitigate the risk of overfitting on the coastal dataset, an early stopping mechanism was employed, which halted the training process if the validation loss did not improve for 15 consecutive epochs.

*Object Detection Training Configuration:* For the spatial localization task, we utilized the Ultralytics PyTorch implementation of the YOLOv8 Nano (YOLOv8n) model. To accelerate convergence and leverage generalized feature extraction, the network was initialized with COCO pre-trained weights (yolov8n.pt). Input images were dynamically resized to a resolution of 640x640 pixels by the native data loader. The model was optimized using the auto-configured Stochastic Gradient Descent (SGD) optimizer with momentum. Consistent with the classification setup, the detection training utilized a batch size of 32, a maximum of 100 epochs, and an early stopping patience of 15 epochs. YOLOv8 was trained using the weighted composite loss objective shown in Equation (3):

$$\text{Total Loss} = \lambda_1 L_{\text{cls}} + \lambda_2 L_{\text{box}} + \lambda_3 L_{\text{dfl}} \quad (3)$$

where $L_{\text{cls}}$is the classification loss, $L_{\text{box}}$is the bounding-box regression loss, $L_{\text{dfl}}$is the Distribution Focal Loss, and $\lambda_1$, $\lambda_2$, and $\lambda_3$are weighting coefficients.

*Wavelet Configuration Details:* To ensure optimal feature extraction across different tasks and architectures, the Discrete Wavelet Transform (DWT) parameters were specifically tailored. For the object detection task using YOLOv8, a first-level decomposition (Level 1) utilizing the Daubechies 4 (db4) wavelet was applied to preserve precise spatial boundaries for bounding box regression. In the classification tasks, a deeper second-level decomposition (Level 2) was utilized to capture broader texture gradients. Within these classification models, the Single-Stream architectures employed the Daubechies 2 (db2) wavelet for efficient high-frequency extraction, whereas the Dual-Stream architectures utilized the db4 wavelet to extract more complex and localized frequency components. Table 1 summarizes the DWT configuration used for each task and input representation.

*Table 1. DWT configuration summary for classification and detection experiments.*

| **Task** | **Model / representation** | **Wavelet family** | **Decomposition level** | **Input channels** |
|---|---|---|---|---|
| Binary classification | Single-Stream wavelet-only (SS-Wavelet-4) | Daubechies 2 (db2) | Level 2 | LL, LH, HL, HH |
| Binary classification | Single-Stream early fusion (SS-RGBW) | Daubechies 2 (db2) | Level 2 | RGB + wavelet-energy map (W) |
| Binary classification | Single-Stream channel replacement (SS-WGB) | Daubechies 2 (db2) | Level 2 | W, G, B |
| Binary classification | Dual-Stream fusion (DS-W4/RGB3) | Daubechies 4 (db4) | Level 2 | RGB stream + LL, LH, HL, HH spectral stream |
| Object detection | YOLOv8n channel replacement (YOLOv8n-WGB) | Daubechies 4 (db4) | Level 1 | W, G, B |
| Object detection | YOLOv8n image fusion / wavelet variants | Daubechies 4 (db4) | Level 1 | Fused RGB-W or wavelet-derived 3-channel inputs |

### 5.2 Evaluation Metrics

To rigorously assess the performance of the proposed methodologies, standard metrics were utilized across both tasks. For the binary classification models, performance was evaluated using Accuracy, Precision, Recall, and the F1-Score. Given the safety-critical nature of rip current monitoring-where failing to detect a hazard poses severe risks to human life-particular emphasis was placed on the Recall metric to ensure the minimization of false negatives.

For the object detection models, overall spatial performance was measured using the mean Average Precision at an Intersection over Union (IoU) threshold of 50% (mAP@50), as well as the average mAP across the 50% to 95% thresholds (mAP@50-95). In addition to the mAP metrics, explicit bounding box Precision, Recall, and the F1-Score were reported. The F1-Score provides a comprehensive understanding of the model's operational reliability and its ability to balance correct detections against false alarms in noisy marine environments.

## 6. Results

This section presents a comprehensive evaluation of the developed models. The performance of the proposed spectral-spatial fusion strategies is quantified across binary classification and object detection tasks, followed by an analysis of computational efficiency and model interpretability. To ensure statistical robustness and mitigate initialization variance, all reported metrics for the classification models represent the average performance across four fixed random seeds (42, 4, 20, and 123), accompanied by their respective standard deviations. The YOLOv8 detection results are reported for the trained configurations evaluated on the held-out test set; seed-averaged variance is not reported for the detection experiments.

### 6.1 Classification Results

Table 2 presents the quantitative evaluation of the tested classification architectures. The standard single-stream RGB baseline (SS-RGB) achieved an average accuracy of 92.31% ± 1.35% and an F1-score of 94.67% ± 1.09%. Incorporating wavelet-derived spectral information generally improved performance when the spectral features were combined with RGB information in an appropriate manner. In contrast, the wavelet-only single-stream model (SS-Wavelet-4) performed below the RGB baseline, indicating that spectral texture alone is insufficient and that visual-spatial information remains important for rip-current classification.

**Table 2. Classification performance of the evaluated RGB, wavelet, channel-replacement, and dual-stream architectures. Values are reported as mean ± standard deviation.**

| Model | Accuracy | F1 | Precision | Recall |
|---|---|---|---|---|
| SS-RGB | 92.31% ± 1.35% | 94.67% ± 1.09% | 92.64% ± 3.73% | 96.79% ± 1.05% |
| SS-Wavelet-4 | 86.78% ± 2.66% | 90.76% ± 1.98% | 87.59% ± 2.87% | 94.23% ± 1.19% |
| SS-RGBW | 91.30% ± 1.53% | 93.88% ± 1.08% | 90.86% ± 1.16% | 97.11% ± 1.32% |
| SS-WGB | 94.48% ± 0.49% | 96.02% ± 0.34% | 95.27% ± 0.67% | 96.79% ± 0.01% |
| **DS-W4/RGB3** | **95.37% ± 1.48%** | **96.66% ± 1.08%** | **95.75% ± 1.01%** | **97.59% ± 1.59%** |
| DS-W3/GB | 94.27% ± 0.82% | 95.91% ± 0.69% | 94.72% ± 1.66% | 96.95% ± 1.59% |
| DS-W4/GB | 93.06% ± 2.47% | 95.20% ± 1.68% | 92.88% ± 3.73% | 97.59% ± 1.05% |
| DS-W3/RGB3 | 92.28% ± 3.02% | 94.57% ± 2.04% | 91.93% ± 3.06% | 97.43% ± 1.63% |

**Note**. SS = single-stream model; DS = dual-stream model; W = wavelet channels; RGB = red-green-blue channels; GB = green-blue channels; RGBW = early-fusion input composed of RGB channels and the wavelet-energy map; WGB = channel-replacement input composed of the wavelet-energy map, green channel, and blue channel. DWT = discrete wavelet transform. Values are reported as mean ± standard deviation across four fixed random seeds where available.

The best classification performance was obtained by the dual-stream model using four wavelet channels and three RGB channels (DS-W4/RGB3), which achieved the highest accuracy (95.37% ± 1.48%) and F1-score (96.66% ± 1.08%). This result suggests that processing spatial RGB information and spectral wavelet information in separate branches before fusion provides a richer representation than using either modality alone. The strong recall of this model (97.59% ± 1.59%) is particularly important for rip-current monitoring, where missed detections represent the most safety-critical error type. In operational terms, high recall directly corresponds to fewer missed hazardous rip-current events, which is critical in a monitoring system designed for beach safety.

The channel-replacement configuration (SS-WGB) also demonstrated strong performance, achieving 94.48% ± 0.49% accuracy and 96.02% ± 0.34% F1-score, outperforming the RGB baseline while preserving a standard three-channel input format. This indicates that replacing the red channel with a wavelet-energy map can inject useful hydrodynamic texture information into a lightweight model without requiring a dual-stream architecture. By contrast, the lower performance of the wavelet-only model suggests that wavelet features are most effective when used as a complement to visual image information rather than as a complete replacement for RGB-based spatial cues.

.

### 6.2 Object Detection Results

Table 3 presents the object detection performance of the evaluated YOLOv8 configurations. The baseline YOLOv8 nano model using standard RGB imagery (YOLOv8n-RGB) achieved 87.30% mAP@50, 41.80% mAP@50-95, and an F1-score of 81.60%. Compared with this baseline, most wavelet-based input representations improved detection performance, indicating that spectral texture information can support more accurate localization of rip-current regions.

**Table 3. Object detection performance of YOLOv8 configurations using RGB, wavelet, image-fusion, channel-replacement, and HSW inputs.**

| Model | mAP@50 | mAP@50-95 | Precision | Recall | F1 | FPS |
|---|---|---|---|---|---|---|
| YOLOv8n-RGB | 87.3% | 41.8% | 81.2% | 89.1% | 81.6% | **74.75** |
| YOLOv8s-RGB | 91.32% | 44.8% | 88.44% | 83.36% | 85.82% | 38.19 |
| YOLOv8n-IF | 91.7% | **47.7%** | 89.7% | 85.9% | 87.76% | 66.72 |
| YOLOv8n-WGB | **94%** | 45.45% | **90.83%** | **90.77%** | **90.5%** | 71.4 |
| YOLOv8n-W | 92.56% | 43.8% | 84.8% | 89.1% | 86.9% | 65.06 |
| YOLOv8n-HSW | 86.7% | 42.4% | 86.59% | 80.77% | 83.57% | 50.93 |

Note. YOLOv8n = YOLOv8 nano; YOLOv8s = YOLOv8 small; RGB = standard red-green-blue input; IF = image fusion; WGB = channel-replacement input composed of the wavelet-

energy map, green channel, and blue channel; W = wavelet-based input; HSW = hue-saturation-wavelet input representation. mAP@50 denotes mean average precision at an IoU threshold of 0.50, and mAP@50-95 denotes mean average precision averaged across IoU thresholds from 0.50 to 0.95. FPS = frames per second.

The strongest overall detection performance was achieved by the channel-replacement configuration (YOLOv8n-WGB), which reached 94.00% mAP@50, 90.83% precision, 90.77% recall, and an F1-score of 90.50%. This result suggests that replacing the red channel with a wavelet-energy map provides YOLOv8 with useful hydrodynamic texture cues while preserving the standard three-channel input format. Importantly, this improvement was obtained with only a modest reduction in inference speed relative to the RGB nano baseline (71.40 FPS compared with 74.75 FPS), indicating that the channel-replacement strategy remains computationally efficient.

The image-fusion configuration (YOLOv8n-IF) achieved the highest mAP@50-95 value (47.70%), suggesting that direct spatial-spectral blending may improve localization quality at stricter IoU thresholds. However, its overall F1-score and recall were lower than those of YOLOv8n-WGB. The larger YOLOv8 small model (YOLOv8s-RGB) improved over the RGB nano baseline but reduced inference speed substantially, from 74.75 FPS to 38.19 FPS. In contrast, the wavelet-enhanced nano configurations preserved real-time processing rates while improving detection accuracy, making them more suitable for UAV-based coastal monitoring workflows. Maintaining approximately 70 FPS on YOLOv8n-WGB suggests feasibility for near-real-time UAV patrols over crowded beaches with typical embedded hardware.

### 6.3 Ablation Studies and Hyperparameter Optimization

While the main results established the superiority of the spectral-spatial fusion architectures, the performance of these models is highly sensitive to specific design choices. To ensure our configurations were robust and data-driven, we conducted a series of ablation studies and hyperparameter evaluations. This section analyzes how variations in terrestrial noise suppression (cropping), wavelet mother functions (db types), and wavelet decomposition levels impact the predictive capabilities across both the classification and detection tasks.

#### *6.3.1 Classification Parameters (Crop Percentage and Wavelet Function)*

To maximize the classification accuracy of both the Dual Stream and Channel Replacement models, we systematically evaluated two critical preprocessing parameters: the spectral resolution of the wavelet transform and the spatial bounding of terrestrial noise.

Influence of Wavelet Mother Function: To determine the optimal spectral representation, we evaluated both architectures across various Daubechies (db) orders. As shown in Fig. 4, the Channel Replacement strategy performs best with db2 (94.48% accuracy), suggesting that lower-order wavelets, which have shorter support, better preserve fine-grained edge gradients required for single-stream fusion. In contrast, the Dual Stream model achieves its peak performance with db4 (95.37% accuracy). This indicates that the parallel processing stream benefits from the improved frequency localization and smoother approximation provided by mid-order wavelets, which helps the attention mechanism focus on relevant hydrodynamic patterns.

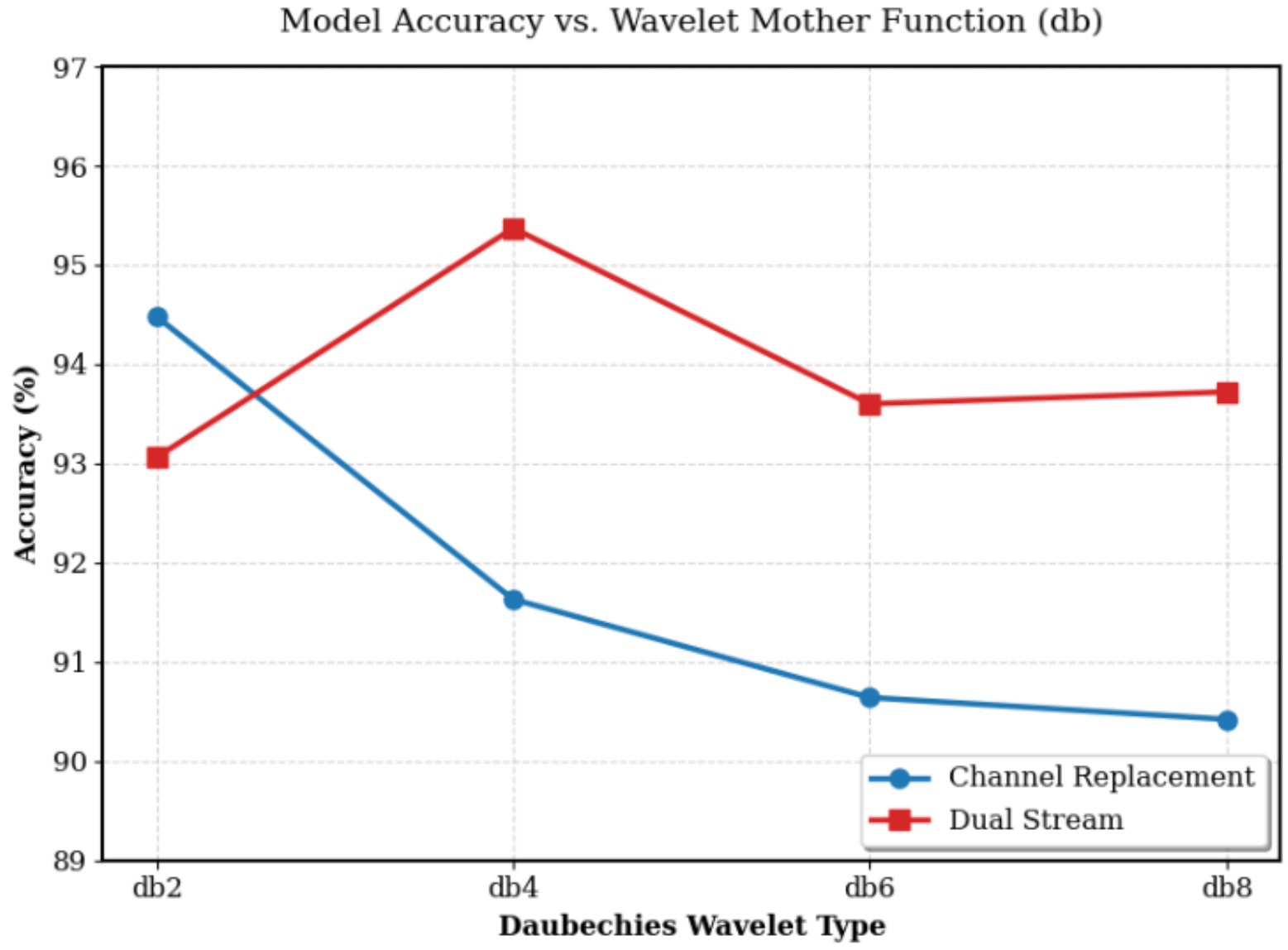

**Fig. 4.** Accuracy comparison of the Channel Replacement and Dual Stream architectures across Daubechies (db) wavelet orders. Data points represent average accuracy across four fixed random seeds.

Impact of Terrestrial Noise Suppression: We further investigated the effect of terrestrial artifacts on model performance by varying the bottom-image crop percentage. Fig. 5 reveals a consistent performance peak for both models at the 20% crop threshold. This trend confirms that removing the bottom fifth of the frame, where sand and shore structures typically reside, successfully eliminates non-hydrodynamic noise that would otherwise interfere with the wavelet-extracted features. However, increasing the crop to 30% leads to a performance drop, likely due to the exclusion of critical near-shore surf zone data, indicating that a 20% crop provides the optimal balance between noise suppression and feature preservation.

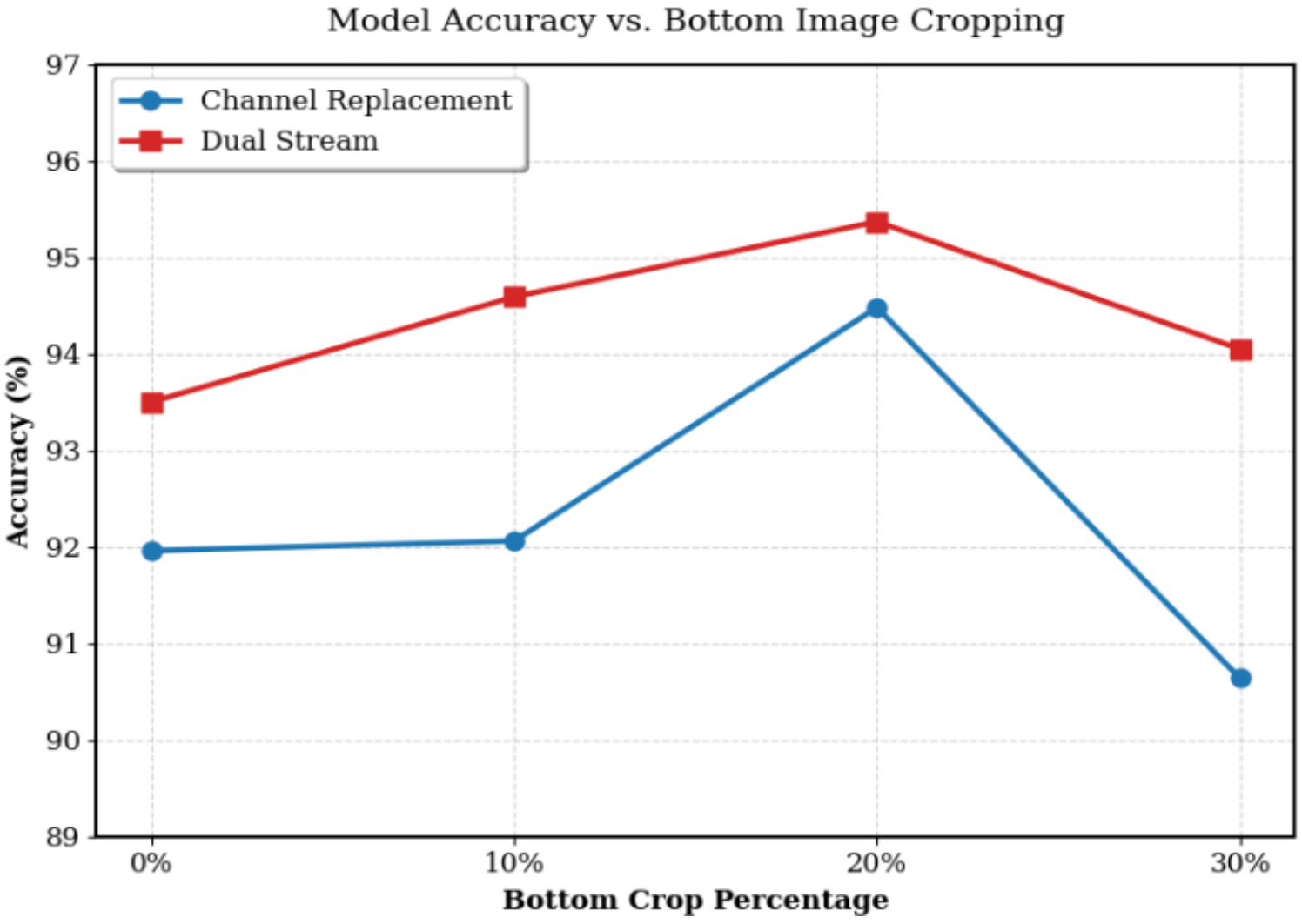

**Fig. 5.** Model accuracy as a function of bottom-image crop percentage. Removing the bottom 20% of the image, corresponding to the terrestrial zone, yields optimal performance for both architectures.

*6.3.2 Detection Parameters (Wavelet Decomposition Levels)*

In addition to the classification parameters, we evaluated the optimal depth of frequency separation for the object detection task by comparing 1-level and 2-level 2D-DWT decompositions. As illustrated in Fig. 6, deeper spectral decomposition (Level 2) degrades detection performance across key evaluated metrics for both the Channel Replacement and Image Fusion architectures.

This trend highlights a fundamental distinction between classification and localization tasks. While our earlier results (Section 6.1) demonstrated that Level 2 decomposition is highly effective for whole-image classification, where capturing global spectral energy distributions is paramount-this deeper decomposition proves suboptimal for object detection. Applying a second level of DWT inherently involves further down sampling, which excessively compresses the spatial dimensions of the feature maps. Because bounding box precision is strictly dependent on high spatial resolution, this loss of spatial granularity hinders the model's ability to draw tight, accurate bounding boxes around rip currents, leading to a drop in mAP scores. Consequently, we conclude that a single-level decomposition (Level 1) provides the optimal balance between spectral feature extraction and spatial preservation for rip current detection.

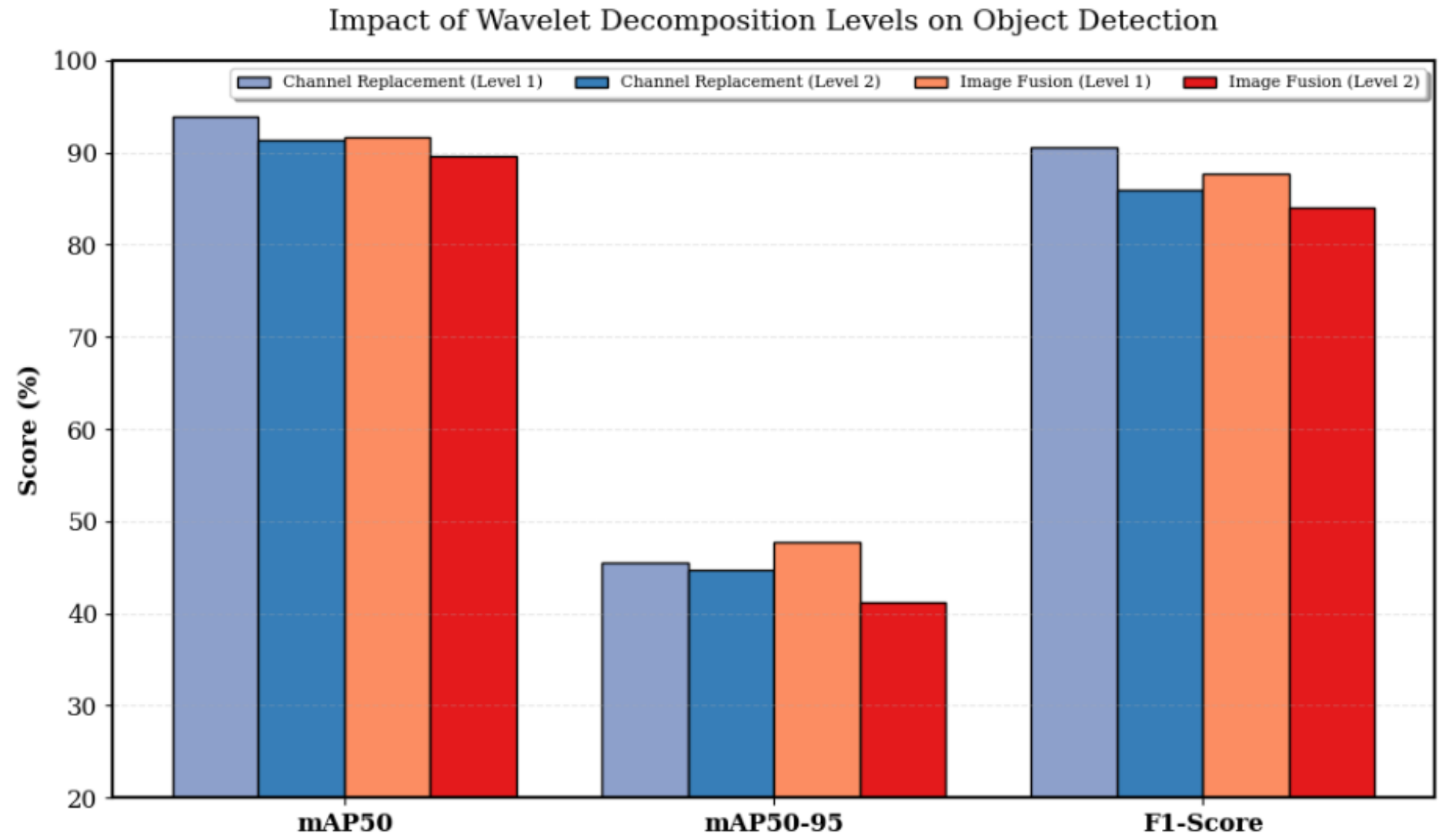

**Fig. 6.** Object detection performance comparison between Level 1 and Level 2 wavelet decompositions across Channel Replacement and Image Fusion configurations.

## 6.4 Explainable AI (XAI) and Failure Mode Analysis

To examine the interpretability of the proposed models for safety-critical coastal monitoring, Explainable AI (XAI) techniques were applied. This subsection provides a visual interpretation of the models' activation patterns, assessing whether their predictions are associated with physically plausible hydrodynamic features rather than obvious environmental noise. These analyses provide qualitative interpretability evidence that helps contextualize the model outputs alongside the quantitative results.

***6.4.1** Weakly Supervised Localization in Classification (Grad-CAM++)*

To interpret the decision-making process of the Dual-Stream classification network, we applied the Grad-CAM++ algorithm to extract spatial class activation maps (CAMs). Unlike the standard Grad-CAM, the Grad-CAM++ variant utilizes second and third-order derivatives of the logits with respect to the final convolutional feature maps, enabling more precise spatial localization and better capturing of object boundaries in complex backgrounds.

Visual analysis suggested that the network's activation patterns varied based on scene contrast. In high-contrast scenarios, the model often activates on the rip current channel itself (Direct Detection). However, in low-contrast conditions where the current is visually ambiguous, the model tends to concentrate activation on breaking surf surrounding the rip, consistent with the identification of the hazard by the distinct "gap" in the wave line (Gap Detection).

To explore this learned behavior for weakly supervised localization, we developed an Adaptive Post-Processing algorithm. To automate the inversion logic and bounding box selection, we applied a center-bias heuristic, predicated on the assumption that in our patch-based classification dataset, the rip current hazard is generally situated near the center of the frame.

The algorithm uses the activation intensity in this center region of the CAM to guide the post-processing step. If the center region exhibits low activation, consistent with Gap Detection, the algorithm inverts the activation map to highlight the negative space, corresponding to the candidate rip-current region. Finally, standard contour detection is applied to the adaptively corrected mask to estimate a bounding box around the hazard. This approach enables the classification network to provide approximate spatial localization cues without requiring bounding-box annotations during training. Representative Grad-CAM++ localization outputs are shown in Fig. 7.

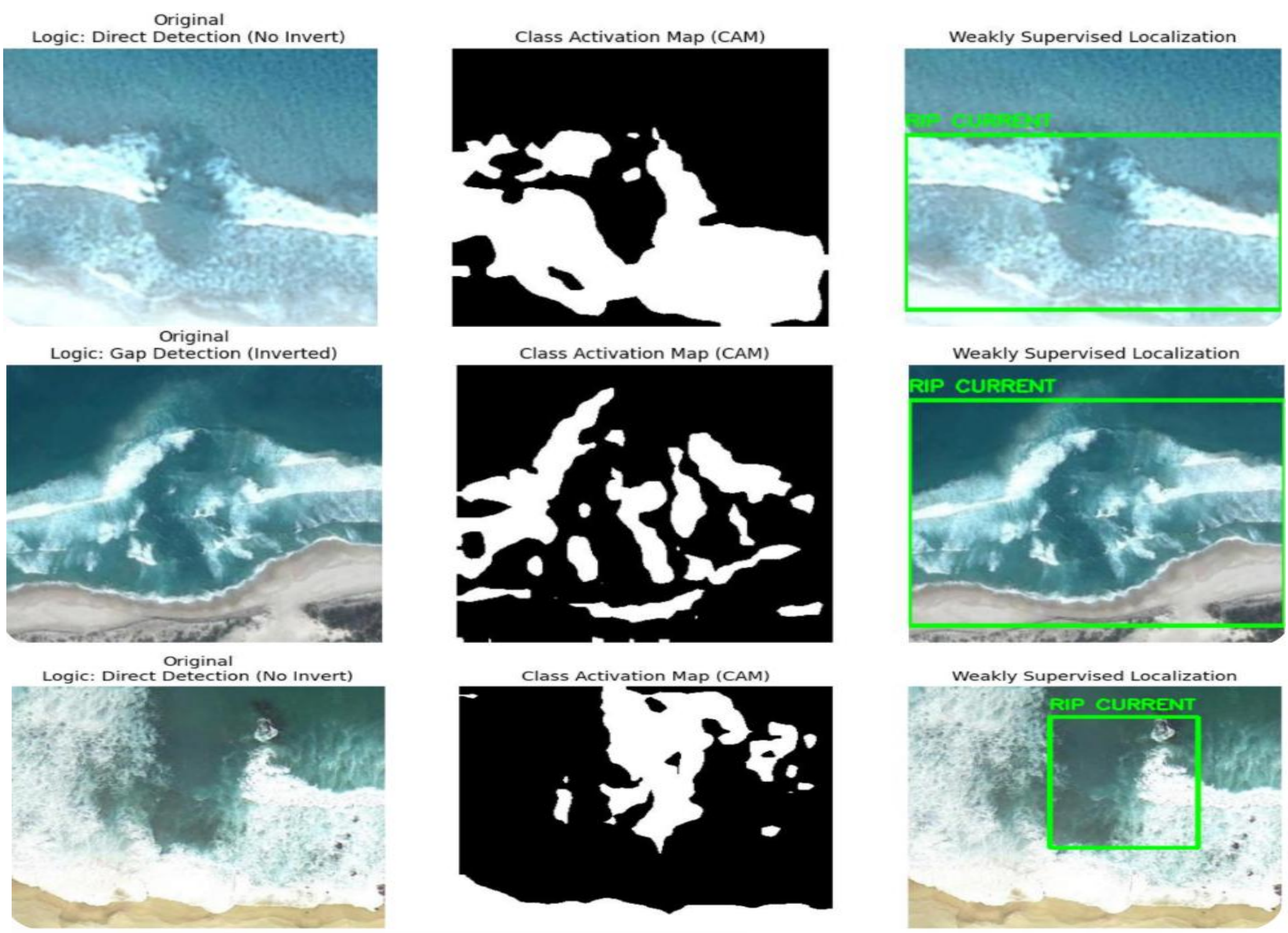

**Fig. 7.** Weakly supervised localization of rip currents using the adaptive Grad-CAM++ algorithm applied to the Dual-Stream classification network.

*6.4.2 Feature Interpretation in Object Detection (EigenCAM)*

To gain insight into the feature representations learned by the YOLOv8 detection model, particularly its utilization of the Channel-Replaced input, we employed the EigenCAM visualization technique. Unlike gradient-based XAI methods, EigenCAM can be useful for object detection architectures as it computes principal components from the 2D activations of the convolutional layers, visualizing the most dominant spatial features without relying on class-specific backpropagation.

The visualization algorithm was applied to the deepest convolutional layer preceding the detection head. As illustrated in Fig. 8, the analysis compares the Channel-Replaced input composite (W,G,B) alongside the generated spatial heatmap and the final bounding box prediction. The resulting EigenCAM heatmaps suggest that the detection model often concentrates activation near the annotated rip-current region.

In several examples, the highest activation zones appear to align with regions where the superimposed high-frequency wavelet energy changes, such as the borders between turbulent breaking waves and the relatively calm water of the rip channel. This suggests that embedding spectral texture into the detection backbone may encourage the network to prioritize hydrodynamically relevant regions rather than relying only on color gradients.

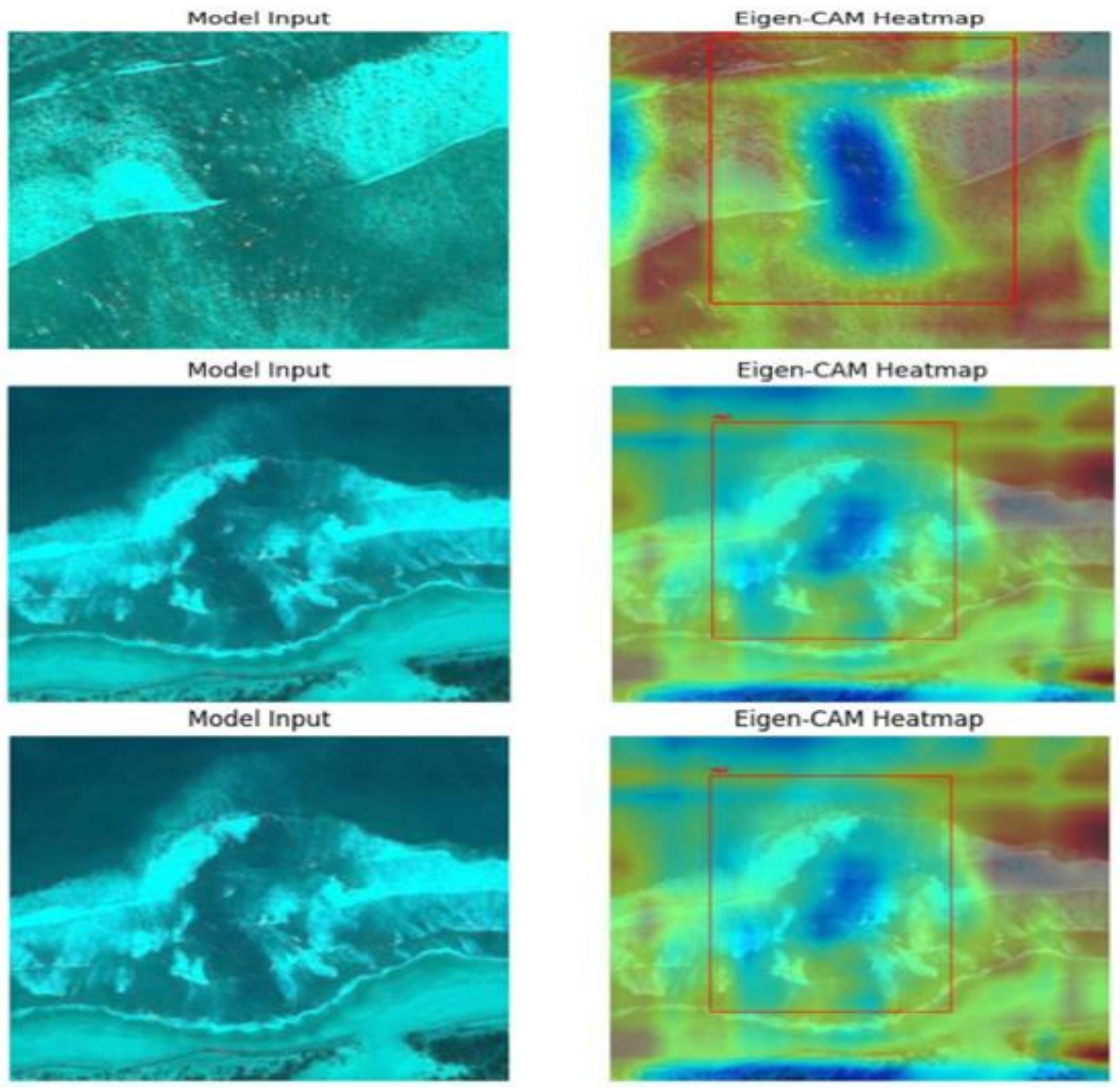

**Fig. 8.** EigenCAM feature visualization of the YOLOv8 detection model, showing the original image, Channel-Replaced W-G-B input, activation heatmap, and predicted rip-current localization.

*6.4.3 Failure Mode Analysis*

While the EigenCAM visualizations suggest that the model learns visually plausible representations related to rip-current indicators, analyzing its misclassifications provides critical insights into the limitations of the current architecture. By examining the activation maps of false positives and false negatives, we can identify the specific oceanographic and textural conditions that confound the network.

False Positives via Bathymetric Mimicry: The integration of high-frequency wavelet components enhances the model's sensitivity to the intensity gradients at the interface between turbulent foam and quiescent water. Although this mechanism is useful for identifying annotated rip currents, it occasionally induces a specific failure mode. The network can misidentify bathymetric troughs or unbroken wave regions as rip channels, prioritizing local textural contrast over the global geometric continuity of the wave surf zone. Examples of this false-positive mode are shown in Fig. 9.

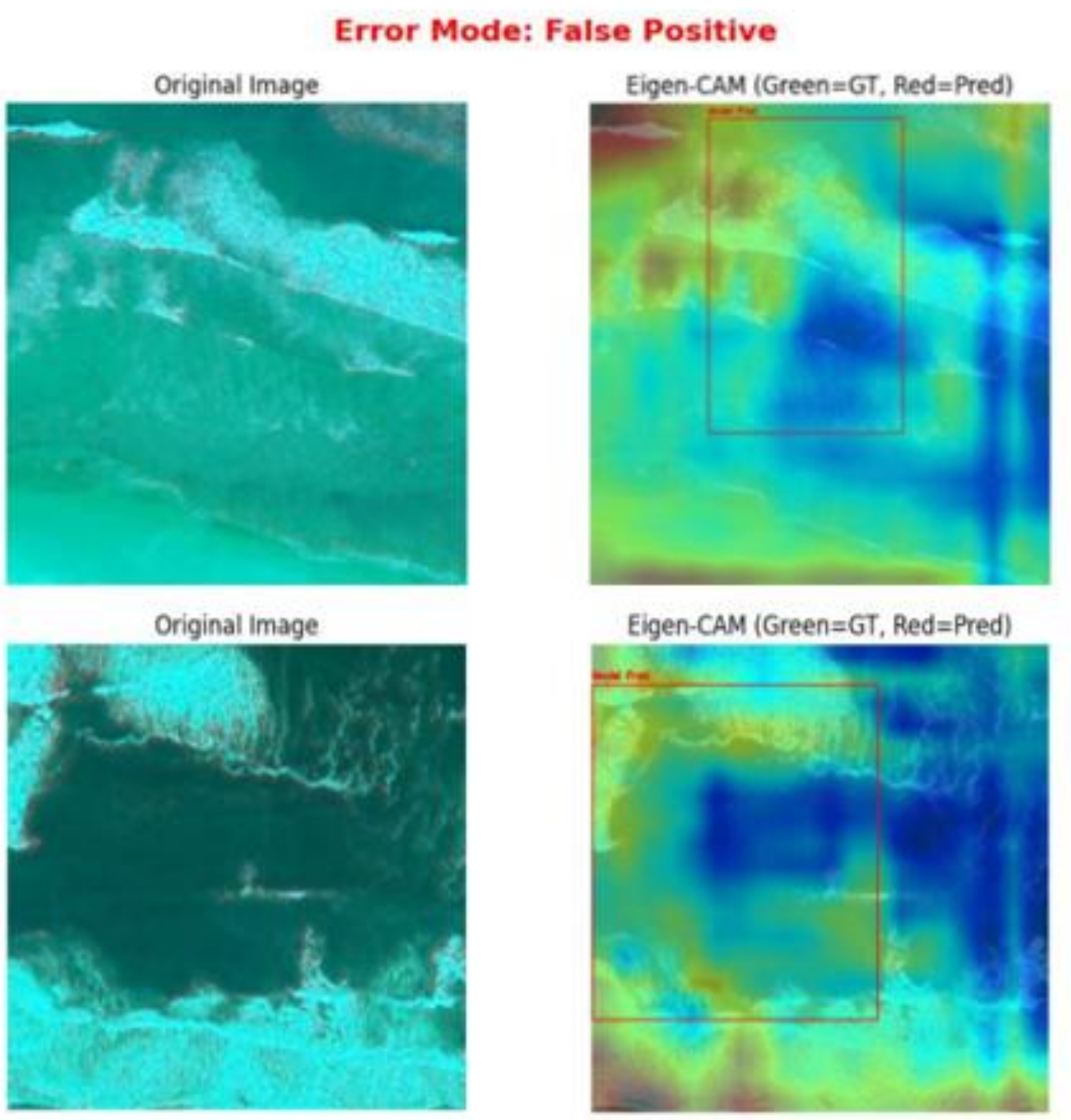

**Fig. 9.** False-positive examples caused by bathymetric mimicry. EigenCAM heatmaps indicate that the model can overemphasize textural contrast in non-rip surf-zone structures.

*False Negatives via Subtle Geometry and Macro-Structural Masking:* Beyond false positive activations, analyzing the model's false negatives (missed detections) reveals specific physical and algorithmic limitations. Object detection architectures, including YOLO, inherently struggle with small-scale targets. While our wavelet-fusion model achieves strong performance on typical, broad rip channels, it occasionally fails when the hazard presents as a highly localized or subtle disruption, such as a minute gap or a narrow wave break.

When these high-resolution coastal images are processed at the network's standardized 640x640 input dimension, these subtle geometric features occupy a highly restricted pixel footprint. Consequently, their weak spatial signature falls below the detection threshold. This resolution challenge is further complicated by a phenomenon termed macro-structural masking. The Wavelet transform can strongly amplify the continuous edges of the surrounding wave fronts. As suggested by the EigenCAM visualizations (Fig. 10), these dominant spectral structures generate strong activations that may mask the localized, low-energy interruptions associated with small rip gaps. As a result, the model focuses primarily on the prominent wave edges and may fail to regress a bounding box over the subtle hazard.

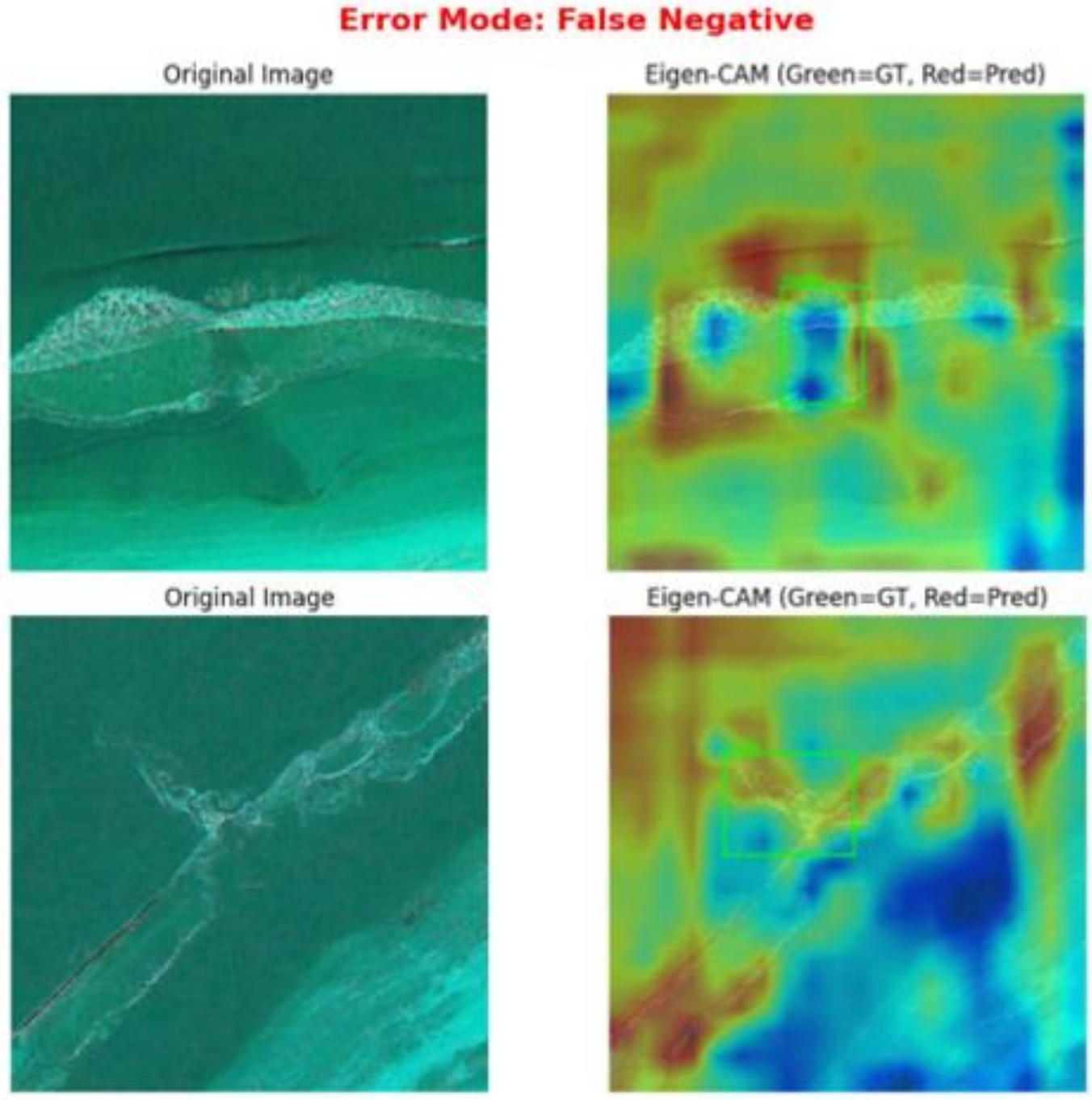

**Fig. 10.** False-negative examples caused by subtle rip-current geometry and macro-structural masking, where dominant wave-front activations obscure small rip gaps.

## 7. Discussion and Conclusions

*7.1 Implications for coastal hazard monitoring*

From a coastal monitoring perspective, the proposed workflow is best understood as a decision-support layer for early identification of hazardous surf-zone conditions. UAV imagery processed with physically informed spectral-spatial representations could support patrol screening, help lifeguards prioritize ambiguous surf-zone regions for closer inspection, and provide coastal managers with interpretable visual cues for beach-safety risk mitigation.

Operational use should remain human-supervised. The method can highlight suspected rip-current zones, but it should not replace lifeguard judgment, local knowledge, real-time sea-state assessment, or site-specific validation. The failure modes identified in this study, including bathymetric mimicry and missed narrow rip-current gaps, indicate that reliability limitations must be explicitly considered before deployment in safety-critical beach management.

*7.2 Main findings and reliability considerations*

This study demonstrates how UAV imagery can be transformed into candidate coastal environmental monitoring outputs by integrating physically meaningful spectral texture information with spatial image features. The proposed framework targets the operational problem of detecting hazardous surf-zone conditions from low-altitude coastal observations. The experimental results show that DWT-derived spectral-spatial information improves both rip-current hazard classification and localization, indicating that wavelet representations can enhance detection of hydrodynamic wave-gap structures that are relevant to beach-safety risk mitigation.

One of the most prominent hazard-monitoring outcomes is the effectiveness of the Channel Replacement strategy, in which the structurally limited Red (R) channel is replaced by a wavelet-based energy map. This approach preserves compatibility with standard 3-channel architectures while delivering substantial performance gains-achieving approximately 94%

mAP@50 in object detection. For UAV-based coastal hazard monitoring, this is important because it embeds a physically interpretable hydrodynamic descriptor into a lightweight detector without requiring a specialized multispectral sensor or major computational overhead. These improvements show that the strategy is a practical way to guide hazard detection toward hydrodynamic boundaries that matter for operational beach safety.

In parallel, the Dual-Stream architecture demonstrates the value of treating UAV coastal imagery as a coupled spatial-spectral observation of hazardous surf-zone conditions. By processing spatial and spectral domains independently before fusion, the model achieved the highest overall performance for classification (exceeding 95% accuracy). Its ability to combine complementary representations-hydrodynamic textures at multiple scales and visual appearance cues-produces a richer feature space that improves both precision and recall. This result suggests that physically informed feature fusion can improve robustness under variable illumination, surf-zone texture, and background clutter, which are central constraints in practical beach-safety monitoring.

Explainable AI (XAI) analyses further support the operational interpretability of the framework. Grad-CAM++ visualizations suggest that the Dual-Stream classifier adaptively switches between two modes of reasoning: Direct Detection, focusing on the rip channel itself in high-contrast scenarios, and Gap Detection, relying on the characteristic break in the surf line when the rip is visually ambiguous. Similarly, EigenCAM visualizations for the detection model indicate that the network concentrates on the transition zones between turbulent breaking waves and calm rip channels. These observations indicate that the models respond to physically meaningful surf-zone patterns rather than arbitrary visual correlations, which is essential if automated UAV products are to be trusted as decision-support tools in coastal monitoring.

## 8. Limitations and Future Work

Although the proposed spectral–spatial framework achieves strong performance, several limitations remain. First, the study relies solely on still images, without incorporating temporal dynamics that often reveal the evolving structure of rip currents. This limits the model's ability to capture flow-driven cues that emerge only across multiple frames. In addition, the analysis focuses on a single spectral representation-first-level Daubechies wavelets-while alternative wavelet families or deeper multi-scale decompositions may provide richer hydrodynamic descriptors. The evaluation also centered on lightweight architectures (task-specific convolutional neural networks and YOLOv8n), leaving open the question of how well the proposed fusion strategies scale to larger, transformer-based, or video-oriented models. External validation across additional beaches, UAV altitudes, illumination regimes, seasons, and sea-state conditions is required before operational deployment. The present study does not establish universal rip-current detectability across all beaches or sea states. It also does not validate detections against direct current measurements. Therefore, the results should be viewed as evidence that wavelet-derived texture features can improve visual detection of annotated rip-current indicators in UAV imagery, rather than as proof of operational readiness for autonomous beach-hazard warning.

Future work may extend the framework through temporal fusion, such as 3D CNNs or transformer-based video models, enabling the detection of subtle or transient rip signatures. Exploring broader spectral approaches, such as Wavelet Packets or Scattering Networks, may further improve robustness under challenging surf conditions. From a coastal environmental monitoring perspective, field deployment on real UAV platforms is especially important, as operational performance must be tested under camera motion, changing altitude, oblique viewing geometry, sun glint, beach crowding, and rapidly evolving wave conditions. Finally, evaluating the method across additional coastal datasets or sensing modalities could provide deeper insight into generalization and support the development of real-time beach-safety risk mitigation systems. Building on the near-duplicate analysis, which supports the independence of the held-out split, future work should further strengthen generalization assessment through validation across independent beaches, UAV altitudes, camera viewpoints, illumination conditions, sea states, and morphodynamic settings.

## Statements and Declarations

### Funding

This research did not receive any specific grant from funding agencies in the public, commercial, or not-for-profit sectors.

### Competing interests

The authors declare that they have no known competing financial interests or personal relationships that could have appeared to influence the work reported in this paper.

### Ethics approval

Not applicable. This study used publicly available imagery and did not involve human participants, human data, or animal subjects.

### Data availability

The original Rip Current Monitoring Dataset (Version 1) used in this study is publicly available through Roboflow Universe at: https://universe.roboflow.com/rip-currents-nso4f/rip-current-monitoring-03cxn. The dataset is distributed under a CC BY 4.0 license and was accessed on 1 June 2026. The source code, preprocessing scripts, wavelet feature-generation pipeline, training configurations, evaluation scripts, and experiment outputs are available at: https://github.com/Yoni2222/RipCurrents-Detection. To ensure long-term reproducibility and persistent access, a static release of this repository has been archived in Zenodo and can be accessed via DOI: 10.5281/zenodo.20480966.

### Author contributions

Yonatan Ben Avraham: Methodology, Software, Validation, Formal analysis, Investigation, Data curation, Visualization, Writing – original draft. Baruch Binyaminov: Methodology, Software, Validation, Formal analysis, Investigation, Data curation, Visualization, Writing – original draft. Yehudit Aperstein: Conceptualization, Methodology, Validation, Supervision, Project administration, Writing – original draft, review and editing. All authors reviewed and approved the final manuscript.